\documentclass[conference,letterpaper]{IEEEtran}

\usepackage[utf8]{inputenc} 
\usepackage[T1]{fontenc}
\usepackage{url}
\usepackage{ifthen}
\usepackage{cite}
\usepackage[cmex10]{amsmath} 

\usepackage{xcolor}
\usepackage{graphicx}
\usepackage{amssymb}
\newtheorem{definition}{Definition}

\newtheorem{proposition}{Proposition}
\usepackage{comment}
\usepackage{algorithm}
\usepackage{algpseudocode}
\usepackage{balance}
\usepackage{booktabs}
\usepackage{subcaption}

\algrenewcommand\algorithmicrequire{\textbf{Input:}}
\algrenewcommand\algorithmicensure{\textbf{Output:}}

\newcounter{algnum}
\begin{document}
\title{DP-MacAdam: Differentially Private Mechanism with Adaptive Clipping and Adaptive Momentum} 

\author{%
 \IEEEauthorblockN{Naima Tasnim, Lalitha Sankar and Oliver Kosut}
\IEEEauthorblockA{
Arizona State University, Tempe, AZ, United States\\
Email: \{ntasnim2, lsankar, okosut\} @ asu.edu}
}


\maketitle

\begin{abstract}
Differentially private stochastic gradient descent (DP-SGD) has become the standard framework for privacy-preserving machine learning, yet its reliance on a fixed gradient clipping threshold to limit sensitivity remains a significant practical limitation. Adaptive clipping algorithms such as AdaClip shift and scale the gradient prior to clipping and adding noise so that the clipped gradient yields a more informative descent direction. The shift and scaling parameters are selected adaptively based on the empirical mean and variance. However, in existing adaptive clipping algorithms, these empirical estimates have not been also used for momentum to accelerate training itself. On the other hand, DP-Adam is an algorithm that exploits Adam-like momentum updates based on the gradient mean and variance to accelerate training, but does not exploit these estimates for adaptive clipping. In this work, we propose Differentially Private Mechanism with Adaptive Clipping and Adaptive Momentum (DP-MacAdam), a novel algorithm that combines these two approaches so as to use the same mean and variance estimates for both clipping and momentum. We perform an analysis showing that DP-MacAdam estimates the gradient variances in a bias-free manner. In addition, we empirically evaluate the privacy and accuracy of DP-MacAdam, demonstrating that it achieves improved model utility compared to DP-SGD, AdaClip, and DP-Adam baselines, without requiring manual tuning of the clipping threshold.
\end{abstract}

\section{Introduction}

The proliferation of machine learning models trained on sensitive data has made privacy-preserving 
optimization a central concern in modern deep learning. Differential privacy (DP)~\cite{dwork2006calibrating} 
has emerged as the gold standard for formal privacy guarantees, providing rigorous, mathematically 
provable protections against the leakage of individual training examples. In the context of deep 
learning, DP is most commonly realized through DP-SGD~\cite{abadi2016deep}, which clips per-sample 
gradients to bound their sensitivity and injects calibrated Gaussian noise before each parameter 
update. While DP-SGD offers strong theoretical guarantees, its empirical performance often lags 
behind modern adaptive optimizers, motivating the development of differentially private variants 
of Adam~\cite{kingma2014adam,bock2018convergence}.

The Adam optimizer maintains exponential moving averages (EMA) of the first and second moments of 
stochastic gradients, yielding coordinate-wise adaptive learning rates. For a variety of ML tasks, including image and natural language processing, Adam consistently outperforms SGD and is now the de facto iterative optimization algorithm for training large models. It is therefore 
natural to apply Adam in the DP setting by feeding privatized gradients directly into its update 
rule, yielding the algorithm DP-Adam \cite{tang2024dp}. 

However, a crucial limitation of both DP-SGD and its Adam-based variants is the 
dependence on a fixed gradient clipping threshold $C$. In practice, the choice of $C$ critically affects the privacy-utility tradeoff: a threshold that is too small discards signal through excessive clipping, while one that is too large increases sensitivity and 
forces larger noise additions. 
Pichapati et al.~\cite{pichapati2019adaclip} address this limitation through AdaClip, a theoretically-motivated adaptive clipping strategy that maintains 
coordinate-wise exponential moving estimates of the gradient mean and variance, using 
them to center, scale, and clip gradients dynamically at each iteration. By adapting the 
clipping geometry to the local gradient distribution, AdaClip provably reduces the 
expected noise added per iteration compared to isotropic $\ell_2$ clipping, yielding 
improved model utility under the same privacy budget.

The two approaches, DP-Adam and AdaClip, address key yet complementary limitations of DP-SGD. 
DP-Adam maintains running estimates of the gradient mean and second moment, and uses them to compute adaptive, coordinate-wise parameter updates. However, DP-Adam still relies on standard clipping, forgoing the noise reduction benefits of adaptive clipping. AdaClip, on the other hand, maintains coordinate-wise estimates of the gradient mean and variance to perform adaptive clipping and reduce the additive noise, but discards these statistics after preprocessing and falls back to a plain SGD update, leaving the adaptive information unused. The combination of adaptive clipping and adaptive momentum updates under differential privacy therefore remains unexplored. 

We propose DP-MacAdam (\textbf{D}ifferentially \textbf{P}rivate \textbf{M}echanism with \textbf{A}daptive \textbf{C}lipping and \textbf{Ada}ptive \textbf{M}omentum), which combines both adaptive clipping and adaptive momentum. Crucially, this combination comes at no additional privacy cost, since all moment estimates are derived entirely from the privatized gradients, and thus, satisfy $(\epsilon,\delta)$-DP by the post-processing property of DP.

\textbf{Main Contributions.} Our main contributions are:
\begin{itemize}
    \item We propose DP-MacAdam, the first algorithm to combine AdaClip-style coordinate-wise adaptive clipping with Adam-style adaptive momentum under differential privacy guarantees.
    \item We derive a novel bias correction factor that yields an unbiased estimate of the gradient variance from the EMA of noise-corrected instantaneous variance estimates, accounting for both DP noise inflation 
    and EMA initialization bias.
    \item We empirically demonstrate that DP-MacAdam outperforms state-of-the-art DP optimizers on MNIST and CIFAR-10 across a range of privacy budgets, without requiring manual tuning of the clipping threshold or dimensionality reduction as a preprocessing step.
\end{itemize}

\textbf{Related Work.} Differentially private stochastic gradient descent was introduced by Abadi et al.~\cite{abadi2016deep}, establishing the canonical recipe of per-sample gradient clipping followed by Gaussian noise addition. Convergence properties of the Adam optimizer were established by Kingma and Ba~\cite{kingma2014adam} and later refined by Bock et al.~\cite{bock2018convergence}. DP-Adam was first introduced in TensorFlow Privacy~\cite{tensorflow_privacy} as a straightforward extension of Adam to the DP setting. Tang et al.~\cite{tang2024dp} identified the DP noise bias in Adam's second moment estimator and proposed DP-AdamBC as a correction. Choi et al.~\cite{choi2025dp} extended this line of work by incorporating decoupled weight decay, yielding DP-AdamW and DP-AdamW-BC with improved empirical performance across image, text, and graph classification tasks. Adaptive clipping strategies for DP-SGD were proposed by Pichapati et al.~\cite{pichapati2019adaclip} through AdaClip, which uses coordinate-wise gradient statistics to minimize the expected noise norm added per iteration. Gilani et al.~\cite{gilani2026geoclip} generalized this idea through GeoClip, which clips and perturbs gradients in a transformed basis, provably reducing added noise compared to isotropic clipping.

\section{Problem Setup}
We briefly review some relevant definitions. Let $\mathcal{D} = \left\{(x^{(i)}, y^{(i)})\right\}_{i=1}^{n}$ denote a dataset of $n$ i.i.d. samples, where $x^{(i)} \in \mathcal{X}$ and $y^{(i)} \in \mathcal{Y}$. Given a parametric model $\theta \in \mathbb{R}^d$ and a loss function $\ell: \mathbb{R}^d \times \mathcal{X} \times \mathcal{Y} \mapsto \mathbb{R}$, the goal is to find the model parameters $\theta$ that minimize the empirical risk $\mathcal{L}(\theta) = \frac{1}{n} \sum_{i=1}^n \ell(\theta; x^{(i)}, y^{(i)})$. 
For notational simplicity, we omit the superscript $(i)$ when referring to a generic sample $x$ and its associated quantities.

Two datasets $\mathcal{D}$ and $\tilde{\mathcal{D}}$ are considered \textit{neighbors}, denoted $\mathcal{D} \sim \tilde{\mathcal{D}}$, if they differ by at most one entry. Differential privacy (DP) is defined with respect to all such neighboring datasets as follows.
\begin{definition}[Differential Privacy~\cite{dwork2006calibrating}]\label{def:dp}
    A randomized algorithm, or mechanism $\mathcal{A}: (\mathcal{X} \times \mathcal{Y})^n \to \mathcal{S}$ is considered ($\epsilon,\delta$)-differentially private (($\epsilon,\delta$)-DP) if, for every pair of neighboring datasets $\mathcal{D}\sim\tilde{\mathcal{D}} \in (\mathcal{X} \times \mathcal{Y})^n$, and for all $S \subseteq \mathcal{S}$, 
    \begin{align}
        \Pr\{ \mathcal{A}(\mathcal{D}) \in S \} \leq e^\epsilon \Pr\{ \mathcal{A}(\tilde{\mathcal{D}}) \in S \} + \delta.
    \end{align}
\end{definition}

\subsection{Private Optimization and Gradient Clipping}
The empirical risk can be minimized using a variety of first-order optimization methods, which differ in how they make use of the gradients $g = \nabla_{\theta}\ell(\theta; x, y)$  to update the model parameters. A standard approach to enforce DP in gradient-based optimization is to bound the sensitivity of each gradient update. Given a per-sample gradient $g_t$ at time step $t$, the $\ell_2$-sensitivity is controlled by clipping:
\begin{align}
    \bar{g}_t = \frac{g_t}{\max\!\left(1, \|g_t\|_2 / C\right)},
\end{align}
which ensures $\|\bar{g}_t\|_2 \leq C$ for a fixed clipping threshold $C > 0$. Gaussian noise $z_t \sim \mathcal{N}(0, \sigma^2 C^2 I_d)$ is then added to the clipped gradient, yielding a privatized estimate $\tilde{g}_t = \bar{g}_t + z_t$ that satisfies $(\epsilon,\delta)$-DP by the Gaussian mechanism~\cite{dwork2006calibrating}.

Using the privatized gradient $\tilde{g}_t$, DP-SGD~\cite{abadi2016deep} updates the model parameters along the negative gradient direction:
\begin{align}
    \theta_t = \theta_{t-1} - \eta \tilde{g}_t,
\end{align}
where $\eta$ is the learning rate. DP-Adam~\cite{tensorflow_privacy} extends the Adam optimizer~\cite{kingma2014adam, bock2018convergence} to the differentially private setting by maintaining exponential moving averages of the privatized gradients and their second moments:
\begin{align}
    m_t &= \beta_1 m_{t-1} + (1-\beta_1)\tilde{g}_t, \quad
    v_t = \beta_2 v_{t-1} + (1-\beta_2)\tilde{g}_t^2,
\end{align}
where $\beta_1, \beta_2$ are exponential decay rates. After computing bias-corrected estimates $\hat{m}_t = m_t / (1-\beta_1^t)$ and $\hat{v}_t = v_t / (1-\beta_2^t)$, DP-Adam updates the model parameters using
\begin{align}
    \theta_t = \theta_{t-1} - \eta \frac{\hat{m}_t}{\sqrt{\hat{v}_t} + \gamma},
\end{align}
where $\gamma$ is a stability constant.

\subsection{Adaptive Clipping}
Both DP-SGD and DP-Adam rely on a fixed, isotropic clipping threshold $C$, applied uniformly across all gradient coordinates to bound the $\ell_2$-sensitivity. As discussed in the introduction, choosing $C$ to avoid over- or under-clipping is a challenge. Pichapati et al. \cite{pichapati2019adaclip} address this through AdaClip, a coordinate-wise adaptive clipping strategy which first centers and scales the gradients using running estimates of the gradient mean $m_t \in \mathbb{R}^d$ and scaling factor $b_t \in \mathbb{R}^d$:
\begin{align}
    w_t = \frac{g_t - m_t}{b_t}, \qquad 
    \bar{w}_t = \frac{w_t}{\max(1, \|w_t\|_2)},
\end{align}
ensuring unit $\ell_2$-norm sensitivity after the transform. Gaussian noise is added to the clipped transformed gradients, and the result is mapped back to the original gradient space:
\begin{align}
    \tilde{g}_t = b_t \odot \left(\bar{w}_t + z_t\right) + m_t. 
\end{align}
The privatized gradient $\tilde{g}_t$ is then used in the gradient descent step.
By centering and scaling gradients coordinate-wise before clipping, AdaClip concentrates the noise budget on dimensions with high variance---where it matters most---and adds less noise to dimensions that carry little information. This is shown to reduce the total noise added per iteration compared to fixed $\ell_2$ clipping, with the advantage growing with the parameter dimension $d$.

\section{Differentially Private Mechanism for Adaptive Clipping with Adaptive Momentum (DP-MacAdam)}\label{sec:dp-macadam}
We present DP-MacAdam in Algorithm~\ref{alg:dp-macadam}, which integrates the coordinate-wise adaptive clipping strategy of AdaClip~\cite{pichapati2019adaclip} with Adam's adaptive momentum updates~\cite{kingma2014adam} under differential privacy. DP-MacAdam has several key properties. First, instead of bounding the noise-corrected instantaneous variance estimate $u_t$ as done in AdaClip, we feed it into an exponential moving average $s_t$ with the same decay rate $\beta_1$ as the mean estimate $m_t$. Second, to ensure that $s_t$ an unbiased estimate of $\tilde{g}_t$, we must account for the bias introduced by the weighted averaging of the past estimates. We derive this bias-correction factor $\kappa_t = 2(\beta_1 - \beta_1^t)/(1 + \beta_1)$ in Appendix~\ref{app:kappa}. To ensure that $\hat{s}_t$ is both lower and upper bounded, we clamp it between constants $h_1$ and $h_2$. Finally, the bias-corrected estimate $\hat{s}_t$ is used to update the adaptive scaling vector $b_t$ following the formula used in AdaClip~\cite{pichapati2019adaclip}. We leave the derivation of the optimal choice of $b_t$ under our EMA-based variance estimate for future work.

\begin{algorithm}[ht]
\caption{DP-MacAdam}
\label{alg:dp-macadam}
{\setlength{\baselineskip}{1.1\baselineskip}
\begin{algorithmic}[1]
\Require Dataset $\mathcal{D}$, learning rate $\eta$, noise multiplier $\sigma$, batch size $B$, hyperparameters $\beta_1, \beta_2, h_1, h_2$, stability constant $\gamma$, number of iterations $T$
\Ensure $\theta_T$
\State \textbf{Initialize:} $\theta_0$; $m_0 = \hat{m}_0 = v_0 = s_0^2 = \mathbf{0},\ b_0 = (1/d) \cdot \mathbf{1}$
\For{$t = 1$ \textbf{to} $T$}
    \State Sample mini-batch $\mathcal{B}_t = \left\{(x_{t,i}, y_{t,i})\right\}_{i=1}^{B}$
    \For{$i = 1$ \textbf{to} $B$}
        \State $g_{t,i} \leftarrow \nabla_\theta \ell\left(\theta_{t-1}; x_{t,i}, y_{t,i}\right)$
        \State $w_{t,i} \leftarrow \dfrac{g_{t,i} - \hat{m}_{t-1}}{b_{t-1}}$
        \State $\bar{w}_{t,i} \leftarrow \dfrac{w_{t,i}}{\max(1, \|w_{t,i}\|_2)}$
    \EndFor
    \State $\tilde{w}_t \leftarrow \dfrac{1}{B}\sum_{i=1}^{B}\bar{w}_{t,i} + z_t$; \quad $z_t \sim \mathcal{N}\!\left(0, \dfrac{\sigma^2}{B^2} I_d\right)$
    \State $\tilde{g}_t \leftarrow b_{t-1} \odot \tilde{w}_t + \hat{m}_{t-1}$
    \State $m_t \leftarrow \beta_1 m_{t-1} + (1 - \beta_1) \tilde{g}_t$
    \State $v_t \leftarrow \beta_2 v_{t-1} + (1 - \beta_2) \tilde{g}_t^2$
    \State $\hat{m}_t \leftarrow \dfrac{m_t}{1 - \beta_1^t}$, \quad $\hat{v}_t \leftarrow \dfrac{v_t}{1 - \beta_2^t}$
    \State $\theta_t \leftarrow \theta_{t-1} - \eta \dfrac{\hat{m}_t}{\sqrt{\hat{v}_t} + \gamma}$
    \State $u_t \leftarrow (\tilde{g}_t - \hat{m}_t) \odot (\tilde{g}_t - \hat{m}_t)$
    \State $s_t \leftarrow \beta_1 s_{t-1} + (1 - \beta_1) u_t$
    \State $c_k^{(t)} \leftarrow \dfrac{(1-\beta_1)\,\beta_1^{\,t-k}}{1-\beta_1^{\,t}}$ for $k = 1, \dots, t$
    \State $\kappa_t \leftarrow (1-\beta_1^t)\Big[1 + \sum_{k=1}^{t} c_k^{(t)}\big(\sum_{i=1}^{k} (c_i^{(k)})^2 - 2c_k^{(k)}\big)\Big]$
    \State $\hat{s}_t \leftarrow \min\!\left(\max\!\left(\dfrac{s_t}{\kappa_t} - b_{t-1}^2 \cdot \dfrac{\sigma^2}{B^2},\ h_1\right), h_2\right)$
    \State $b_t \leftarrow \hat{s}_t^{1/4} \cdot \left(\sum_{j=1}^d \hat{s}_{t,j}^{1/2} \right)^{1/2}$
\EndFor
\State \textbf{return} $\theta_T$
\end{algorithmic}}
\end{algorithm}

\subsection{Privacy Analysis}
The privacy of DP-MacAdam follows that of DP-SGD and DP-Adam. Since the only interaction with the private dataset $\mathcal{D}$ occurs through the per-sample gradients, privacy is guaranteed at the point of noise injection in step 10 of Algorithm~\ref{alg:dp-macadam}. All downstream computations---the moment estimates $m_t$, $v_t$, the variance estimate $s_t$, and the adaptive scaling vector $b_t$---are functions solely of the privatized gradient $\tilde{g}_t$ and public hyperparameters. By the post-processing property of differential privacy~\cite{dwork2006calibrating}, DP-MacAdam therefore inherits the same $(\epsilon, \delta)$-DP guarantee as DP-SGD, for any privacy accountant used to track the cumulative privacy loss across $T$ iterations. We use the Connect-the-Dots~\cite{doroshenko2022connect} privacy accountant in our experiments. We state the privacy guarantee of Algorithm~\ref{alg:dp-macadam} formally in Proposition 1. The proof follows directly from the privacy analysis of DP-SGD~\cite{abadi2016deep}.
\begin{proposition}[Privacy guarantee of DP-MacAdam]
For any privacy accountant $\mathrm{Compose}(T, \theta_{1,\ldots,T})$ under which $\mathrm{DP\text{-}SGD}(\theta, C, \mathcal{D}, \sigma, B)$~\cite{abadi2016deep} satisfies $(\epsilon, \delta)$-DP, $\mathrm{DP\text{-}MacAdam}(\theta, \mathcal{D}, \sigma, B)$ satisfies the same $(\epsilon, \delta)$-DP guarantee under the same accountant.
\end{proposition}

\subsection{Noise Scaling}
In DP-MacAdam, the adaptive parameters $\hat{m}_t$ and $b_t$ naturally set the clipping threshold, so there is no need for a separate clipping hyperparameter $C$, as in DP-SGD and DP-ADAM. Thus, in step 6 of Algorithm~\ref{alg:dp-macadam}, centering and scaling each entry of the gradient vector $g_t$ ensures that the $\ell_2$-norm of $\bar{w}_t$ is at most $1$. Gaussian noise is then added to the average of the clipped transformed gradients $\bar{w}_t$ rather than to each individual gradient. For a batch size  $B$, the average of $B$ clipped gradients has $\ell_2$-sensitivity $1/B$ (each $\bar{w}_t^{(i)}$ has at most unit norm, and changing one sample affects the average by at most $1/B$), so the noise is scaled accordingly as $z_t \sim \mathcal{N}(0, \frac{\sigma^2}{B^2}I_d)$. This is equivalent to adding noise $\mathcal{N}(0, \sigma^2 I_d)$ to each individual gradient and averaging, but is more efficient in practice. After the map-back in step 11, the effective noise variance is $\frac{b_{t-1}^2 \sigma^2}{B^2}$; this is why in step 19, this term is subtracted to calculate $\hat{s}_t$, so as to estimate the variance of the un-noised gradient.\footnote{Clipping introduces a non-linear function to the gradient prior to noise, so even removing the noise does not give a completely unbiased estimate of the pure gradients. The adaptive clipping will mitigate this limitation over the course of the training process.}

\subsection{Bias-Corrected Variance Estimation}\label{sec:kappa}
In Algorithm \ref{alg:dp-macadam}, steps 5--11 follow the AdaClip approach to compute the centered and clipped gradient; steps 12--15 follow the Adam approach of computing the moving average and second moment of the gradients, and then taking the descent step. A key aspect of DP-MacAdam is in combining these statistics to update the center and scale factors for the next batch. This is captured in steps 16--19 where we form an instantaneous estimate of the gradient variance (step 16), compute its moving average (step 17), correct for the bias introduced by the moving average (step 18), bound the estimated variance (step 19), and use AdaClip's update rule for center and scaling parameters for the next batch (step 20).

In Appendix~\ref{app:kappa}, we derive the bias correction factor $\kappa_t$ used in step 19 of Algorithm~\ref{alg:dp-macadam}. The goal is to obtain an unbiased estimate $\hat{s}_t$ of the true gradient variance $\sigma_g^2$ from the exponential moving average (EMA) $s_t$. The challenge here arises from the fact that $s_t$ is a weighted average of past noise-corrected variance estimates $u_i$, each of which is itself centered around a noisy mean $\hat{m}_t$---so the bias correction must account for both the EMA initialization bias and the correlation structure introduced by the weighted averaging. 

It is worth recalling that the empirical estimate of the variance from $n$ samples (with equal weights) requires scaling by $1/(n-1)$ rather than $1/n$ to ensure an unbiased estimate. This same phenomenon must be accounted for here for the exponential weighted average. 

\subsection{Extension to DP-MacAdamBC}
A related line of work addresses a different but complementary bias in DP-Adam. 
Tang et al.~\cite{tang2024dp} observe that the addition of DP noise introduces a constant upward shift $\Phi = (\sigma C / B)^2$ in Adam's second moment estimate $\hat{v}_t$, since the noise is independent of the gradient and its variance adds directly to the EMA of squared gradients. Under typical DP parameters, this bias dominates $\hat{v}_t$, effectively reducing DP-Adam to DP-SGD with momentum and a specific learning rate schedule. They propose DP-AdamBC, which corrects for this by subtracting the known bias $\Phi$ from the second moment estimate in the parameter update:
\begin{align}
    \theta_t = \theta_{t-1} - \eta \, \hat{m}_t / \sqrt{\max(\hat{v}_t - \Phi,\ \gamma')}.
\end{align}
Since $\Phi$ is computable from public hyperparameters $\sigma$, $C$, and $B$, this correction comes at no additional privacy cost. The adaptive clipping strategy of DP-MacAdam is orthogonal to this correction and could be combined with DP-AdamBC by replacing the standard parameter update in Algorithm~\ref{alg:dp-macadam} with the bias-corrected update above, where in place of the constant $\Phi$ we use $(\sigma / B)^2$. We call this variant of the algorithm DP-MacAdamBC.

\section{Experimental Results}
We compare the performance of our proposed DP-MacAdam with that of DP-SGD, AdaClip, DP-Adam, and DP-AdamBC. We focus on image classification tasks with MNIST~\cite{lecun1998mnist} and CIFAR-10~\cite{krizhevsky2009cifar} datasets. For MNIST, we train a two-layer fully connected neural network from scratch. The network takes flattened $28 \times 28$ pixel inputs and passes them through a hidden layer of $1000$ units with ReLU activation, followed by a linear output layer of $10$ units corresponding to the digit classes. The model has a total of $d = 795{,}010$ trainable parameters. For CIFAR-10~\cite{krizhevsky2009cifar}, we use a 5-layer CNN similar to that of \cite{tang2024dp}, trained from scratch. The model takes $32 \times 32$ RGB images as input and has a total of $d = 582{,}346$ trainable parameters.

Throughout our experiments, we use learning rate $\eta=0.1$ for DP-SGD and $\eta=0.001$ for the rest. We use the standard Adam momentum coefficients $\beta_1 = 0.9, \beta_2 =0.999,$ and $\gamma = 10^{-8}$. All models are trained for 5 epochs with a batch size $B=256$ for MNIST and $B=512$ for CIFAR-10, respectively. We adopt the same hyperparameter settings reported by \cite{tang2024dp} and \cite{pichapati2019adaclip} where applicable. The clipping norm for DP-SGD, DP-Adam and DP-AdamBC is set to $C=1.0$. For various noise multiplier $\sigma$, the overall privacy budget $\epsilon$ is computed using the Connect-the-Dots accountant~\cite{doroshenko2022connect}, with the privacy parameter $\delta=10^{-5}$. All results are reported as mean $\pm$ standard deviation over 5 random seeds.
\begin{table}[h]
\centering
\caption{Test accuracy (\%) on MNIST dataset across noise multipliers ($\sigma$); $h_1 = 10^{-9}$ and $h_2 = 10^{-6}$.}
\label{tab:mnist-new}
\resizebox{\columnwidth}{!}{%
\begin{tabular}{lcccccccc}
\hline
Algorithm & $\sigma=0.5$ & $\sigma=0.6$ & $\sigma=0.7$ & $\sigma=0.8$ & $\sigma=0.9$ & $\sigma=1.0$ \\
& $\varepsilon=7.49$ & $\varepsilon=4.00$ & $\varepsilon=2.33$ & $\varepsilon=1.46$ & $\varepsilon=1.02$ & $\varepsilon=0.80$ \\
\hline
DP-SGD        & 90.0 $\pm$ 0.1 & 89.8 $\pm$ 0.1 & 89.1 $\pm$ 0.1 & 88.7 $\pm$ 0.1 & 88.2 $\pm$ 0.1 & 87.0 $\pm$ 0.1 \\
AdaClip       & 86.5 $\pm$ 0.8 & 84.9 $\pm$ 0.9 & 83.4 $\pm$ 1.0 & 81.5 $\pm$ 1.2 & 79.8 $\pm$ 1.7 & 79.8 $\pm$ 1.7 \\
DP-Adam       & 92.8 $\pm$ 0.1 & 92.5 $\pm$ 0.1 & 92.2 $\pm$ 0.1 & 91.9 $\pm$ 0.2 & 91.7 $\pm$ 0.1 & 91.7 $\pm$ 0.1 \\
DP-AdamBC     & 87.9 $\pm$ 0.4 & 86.8 $\pm$ 0.4 & 85.8 $\pm$ 0.5 & 84.8 $\pm$ 0.5 & 84.0 $\pm$ 0.4 & 84.0 $\pm$ 0.4 \\
DP-MacAdam & \textbf{93.1 $\pm$ 0.1} & \textbf{92.8 $\pm$ 0.1} & \textbf{92.6 $\pm$ 0.2} & \textbf{92.4 $\pm$ 0.1} & \textbf{92.2 $\pm$ 0.2} & \textbf{91.9 $\pm$ 0.2} \\
DP-MacAdam-BC & 92.2 $\pm$ 0.1 & 92.3 $\pm$ 0.1 & 92.2 $\pm$ 0.2 & 92.1 $\pm$ 0.2 & 92.0 $\pm$ 0.1 & 91.8 $\pm$ 0.1 \\
\hline
\end{tabular}}
\end{table}
\vspace{-3mm}
\begin{table}[h]
\centering
\caption{Test accuracy (\%) on CIFAR-10 dataset across noise multipliers ($\sigma$); $h_1 = 5 \times 10^{-5}$ and $h_2 = 1.0$.}
\label{tab:cifar10}
\resizebox{\columnwidth}{!}{%
\begin{tabular}{lccccc}
\hline
\textbf{Algorithm} & $\sigma=0.5$ & $\sigma=0.6$ & $\sigma=0.8$ & $\sigma=1.1$ & $\sigma=1.5$ \\
& $\varepsilon=10.40$ & $\varepsilon=5.88$ & $\varepsilon=2.45$ & $\varepsilon=1.11$ & $\varepsilon=0.66$ \\
\hline
DP-SGD               & 43.76 $\pm$ 0.3 & 43.75 $\pm$ 0.3 & 43.67 $\pm$ 0.3 & 43.76 $\pm$ 0.3 & 43.60 $\pm$ 0.3 \\
AdaClip              & 29.13 $\pm$ 0.6 & 19.93 $\pm$ 0.4 & 22.96 $\pm$ 0.5 & 10.01 $\pm$ 0.1 & 10.01 $\pm$ 0.1 \\
DP-Adam              & 58.90 $\pm$ 0.3 & 56.96 $\pm$ 0.2 & 54.59 $\pm$ 0.2 & 52.04 $\pm$ 0.1 & \textbf{49.92 $\pm$ 0.2} \\
DP-AdamBC            & 58.34 $\pm$ 0.2 & 55.98 $\pm$ 0.3 & 53.02 $\pm$ 0.2 & 48.70 $\pm$ 0.6 & 43.42 $\pm$ 0.6 \\
DP-MacAdam           & \textbf{59.43 $\pm$ 0.1} & \textbf{58.12 $\pm$ 0.1} & \textbf{55.57 $\pm$ 0.31} & \textbf{52.38 $\pm$ 0.1} & 48.49 $\pm$ 0.4 \\
DP-MacAdam-BC        & 44.33 $\pm$ 1.4 & 44.06 $\pm$ 2.1 & 39.66 $\pm$ 1.0 & 26.89 $\pm$ 10.5 & 10.07 $\pm$ 0.1 \\
\hline
\end{tabular}}
\end{table}

We observe that DP-MacAdam outperforms the state-of-the-art algorithms on both MNIST and CIFAR-10 across most evaluated privacy budgets, with the exception of the highest noise setting on CIFAR-10 ($\sigma=1.5$). DP-MacAdam-BC, however, underperforms relative to DP-MacAdam on MNIST, and performs poorly on CIFAR-10. Following \cite{tang2024dp}, we hypothesize that the second moment bias correction yields larger gains on tasks where Adam and sign descent outperform SGD in the non-private case. It is worth noting that unlike \cite{pichapati2019adaclip}, DP-MacAdam (i) operates directly on the raw inputs without spending privacy budget on dimensionality reduction via PCA, and (ii) applies clamping only after the bias-corrected variance estimate $\hat{s}_t$ is computed rather than on the instantaneous estimate $u_t$, yielding a more stable $b_t$.

\section{Conclusion}
We have proposed DP-MacAdam, a differentially private optimization algorithm that combines coordinate-wise adaptive clipping from AdaClip with Adam momentum updates. By maintaining running estimates of the gradient mean and variance from the privatized gradients, DP-MacAdam adapts the clipping threshold coordinate-wise at each iteration while simultaneously using the same statistics to drive adaptive parameter updates. We have derived a novel bias correction factor $\kappa_t$ that accounts for both the DP noise inflation and the initialization bias introduced by the exponential moving average of the variance estimate. Empirical results on real-world datasets show that DP-MacAdam is consistently better than DP-Adam and DP-AdamBC for typical privacy values. Future work includes a formal convergence analysis of DP-MacAdam, which can lead to determining the optimal choice of the scaling vector $b_t$ as well as empirical evaluation on more complex learning tasks (e.g., NLP) over large datasets. 

\appendices
\section{Bias Correction Factor $\kappa_t$}\label{app:kappa}
In this analysis, we use $\beta$ in place of $\beta_1$ for simplicity. We also perform our analysis in an element-by-element basis, so we treat each vector as as a scalar.
The EMA of the noisy gradients is
\begin{align}
    m_t = \beta m_{t-1} + (1-\beta) \tilde{g}_t 
    = \sum_{k=1}^t (1-\beta) \beta^{t-k} \tilde{g}_k,
\end{align}
assuming $m_0 = \mathbf{0}$. The bias-corrected mean estimate is thus:
\begin{align}
    \hat{m}_t &= \dfrac{m_t}{1-\beta^t} = \sum_{k=1}^t \dfrac{(1-\beta)\beta^{t-k}}{1-\beta^t} \tilde{g}_k.
\end{align}
Let $c_k^{(t)} = \dfrac{(1-\beta)\beta^{t-k}}{1-\beta^t}$. Note that $\sum_{k=1}^t c_k^{(t)} = 1$. Now, the instantaneous variance of $\tilde{g}_i$ is $u_t = \left(\tilde{g}_t - \hat{m}_t\right)^2.$
Applying EMA, we have
\begin{align}
    s_t &= \beta s_{t-1} + (1-\beta) u_t
    = (1-\beta^t) \sum_{k=1}^{t} c_k^{(t)} u_k.
\end{align}
Taking the expectation, we obtain
\begin{align}
    \mathbb{E}[s_t] = (1-\beta^t) \sum_{k=1}^{t} c_k^{(t)} \mathbb{E}[u_k].
\end{align}
Now, the expectation of $u_k$ at a given step $k \leq t$ is
\begin{align}
    \mathbb{E}[u_k] &= \mathbb{E}\left[(\tilde{g}_k - \hat{m}_k)^2\right]\\
    &= \mathbb{E}\left[\left(\tilde{g}_k \sum_{i=1}^{k} c_i^{(k)} - \sum_{i=1}^{k} c_i^{(k)} \tilde{g}_i\right)^2\right] \\
    &= \sum_{i=1}^{k}\sum_{j=1}^{k} c_i^{(k)} c_j^{(k)}\, \mathbb{E}\left[(\tilde{g}_k - \tilde{g}_i)(\tilde{g}_k - \tilde{g}_j)\right].
\end{align}
Let $\mu_g$ and $\sigma_g^2$ be the true mean and variance of the unnoised clipped gradient. Since $\tilde{g}_k$ has noise added, its mean and variance are $\mu_g$ and 
$\sigma_g^2+\frac{b_{k-1}^2 \sigma^2}{B^2}$. We assume that the clipping parameter $b_k$ changes slowly enough that it is a constant, so $b_k=b$. We also assume that across batches the gradients $g_k$ are i.i.d. 
Now, we  consider different cases:
\begin{itemize}
    \item $i = j = k$, or $i \neq j, i = k$, or $i \neq j, j = k$:
    \begin{align}
        \mathbb{E}\left[(\tilde{g}_k - \tilde{g}_i)(\tilde{g}_k - \tilde{g}_j)\right] = 0.
    \end{align}

    \item $i = j,\ i \neq k,\ j \neq k$:
    \begin{align}
        &\mathbb{E}\left[(\tilde{g}_k - \tilde{g}_i)^2\right]
        = \mathbb{E}[\tilde{g}_k^2] - 2\mathbb{E}[\tilde{g}_k\tilde{g}_i] + \mathbb{E}[\tilde{g}_i^2] \notag \\ 
        &\quad= 2\left(\sigma_g^2 + \frac{b^2 \sigma^2}{B^2} + \mu_g^2 \right) - 2\mu_g^2
        = 2\left(\sigma_g^2+\frac{b^2 \sigma^2}{B^2}\right).
    \end{align}

    \item $i \neq j \neq k$:
    \begin{align}
        \hspace{-.25in}\mathbb{E}\left[(\tilde{g}_k - \tilde{g}_i)(\tilde{g}_k - \tilde{g}_j)\right]
        &= \sigma_g^2+\frac{b^2 \sigma^2}{B^2}.
    \end{align}
\end{itemize}
Combining everything, we have
\begin{align}
    \mathbb{E}[u_k] = \left(2\sum_{i \neq k} \big(c_i^{(k)}\big)^2 + \sum_{i \neq j \neq k} c_i^{(k)} c_j^{(k)}\right)\left(\sigma_g^2+\frac{b^2\sigma^2}{B^2}\right).
\end{align}
Now we take the weighted sum,
\begin{align}
    \sum_{k=1}^{t} c_k^{(t)} \mathbb{E}[u_k]
    &= \sum_{k=1}^{t} c_k^{(t)} \left(2\sum_{i \neq k} \big(c_i^{(k)}\big)^2 + \sum_{i \neq j \neq k} c_i^{(k)} c_j^{(k)}\right) \notag \\ &\qquad \left(\sigma_g^2+\frac{b^2\sigma^2}{B^2}\right)\\
    &= A \left( \sigma_g^2+\frac{b^2\sigma^2}{B^2}\right),
\end{align}
where
\begin{align}
    A &= \sum_{k=1}^{t} c_k^{(t)} \left(2\sum_{i \neq k} \big(c_i^{(k)}\big)^2 + \sum_{i \neq j \neq k} c_i^{(k)} c_j^{(k)}\right) \\
    &= \sum_{k=1}^{t} c_k^{(t)}\left(2\sum_{i \neq k} \big(c_i^{(k)}\big)^2 + \left(\sum_{i \neq k} c_i^{(k)}\right)^2 - \sum_{i \neq k} \big(c_i^{(k)}\big)^2\right) \\
    &= \sum_{k=1}^{t} c_k^{(t)}\left(\sum_{i \neq k} \big(c_i^{(k)}\big)^2 + \big(1 - c_k^{(k)}\big)^2\right) \\
    &= \sum_{k=1}^{t} c_k^{(t)}\left(\sum_{i=1}^{k} \big(c_i^{(k)}\big)^2 - \big(c_k^{(k)}\big)^2 + 1 - 2c_k^{(k)} + \big(c_k^{(k)}\big)^2\right) \\
    &= 1 + \sum_{k=1}^{t} c_k^{(t)}\left(\sum_{i=1}^{k} \big(c_i^{(k)}\big)^2 - 2c_k^{(k)}\right)
\end{align}
Therefore,
\begin{align}
    \mathbb{E}[s_t] &= (1-\beta^t) \sum_{k=1}^{t} c_k^{(t)}\, \mathbb{E}[u_k] \notag \\
    &= (1-\beta^t)\left[1 + \sum_{k=1}^{t} c_k^{(t)}\left(\sum_{i=1}^{k} \big(c_i^{(k)}\big)^2 - 2c_k^{(k)}\right)\right] \notag \\ &\qquad \cdot \left(\sigma_g^2+\frac{b^2\sigma^2}{B^2}\right),
\end{align}
where the bias correction factor is:
\begin{align}
    \kappa_t &= (1-\beta^t)\left[1 + \sum_{k=1}^{t} c_k^{(t)}\left(\sum_{i=1}^{k} \big(c_i^{(k)}\big)^2 - 2c_k^{(k)}\right)\right],
\end{align}
We can also see that $\frac{s_t}{\kappa_t}-b_{t-1}^2\frac{\sigma^2}{B^2}$ is an un-biased estimate of $\sigma_g^2$; this explains the exact form of step 19 in Algorithm~\ref{alg:dp-macadam} prior to clamping.

\section*{Acknowledgment}
We would like to thank Sasmita Sivabalarajan of IISc for noticing a minor error in Appendix \ref{app:kappa}. This work is supported in part by NSF grants CIF-2312666 and SCH-2205080. 


\IEEEtriggeratref{6}
\bibliographystyle{IEEEtran}
\bibliography{bibliography26}

\end{document}